\documentclass[runningheads]{llncs}
\usepackage{graphicx}
\usepackage{array}
\usepackage{xcolor}
\usepackage{caption}
\usepackage{subcaption}
\usepackage{cleveref}
\usepackage{cite}
\usepackage{wrapfig}
\usepackage{multirow}
\begin{document}
\title{Stylometric Detection of AI-Generated Text in Twitter Timelines}

% \titlerunning{Sty.det.mac.gen.text}

\author{Tharindu Kumarage\inst{1} \and
Joshua Garland\inst{1} \and
Amrita Bhattacharjee \inst{1} \and 
Kirill Trapeznikov \inst{2} \and 
Scott Ruston \inst{1} \and
Huan Liu \inst{1}
}
\authorrunning{Kumarage et al.}
\institute{Arizona State University, Tempe AZ, USA \and
STR, Woburn MA, USA \\
\email{\{kskumara,jtgarlan,abhatt43,sruston,huanliu\}@asu.edu \inst{1}}
\email{kirill.trapeznikov@str.us} \inst{2}}

\maketitle              % typeset the header of the contribution
\begin{abstract}

Recent advancements in pre-trained language models have enabled convenient methods for generating human-like text at a large scale. Though these generation capabilities hold great potential for breakthrough applications, it can also be a tool for an adversary to generate misinformation. In particular, social media platforms like Twitter are highly susceptible to AI-generated misinformation. A potential threat scenario is when an adversary hijacks a credible user account and incorporates a natural language generator to generate misinformation. Such threats necessitate automated detectors for AI-generated tweets in a given user's Twitter timeline. However, tweets are inherently short, thus making it difficult for current state-of-the-art pre-trained language model-based detectors to accurately detect at what point the AI starts to generate tweets in a given Twitter timeline. In this paper, we present a novel algorithm using stylometric signals to aid detecting AI-generated tweets. 
% The problem consists of two related tasks. 
We propose models corresponding to quantifying stylistic changes in human and AI tweets in two related tasks: Task 1 - discriminate between human and AI-generated tweets, and Task 2 - detect if and when an AI starts to generate tweets in a given Twitter timeline. Our extensive experiments demonstrate that the stylometric features are effective in augmenting the state-of-the-art AI-generated text detectors. 

\end{abstract}
\section{Introduction}

With the recent advances in transformer-based language models, we see tremendous improvements in  natural language generation (NLG). Consequently, with the proliferation of pre-trained language  models (PLM) such as Grover \cite{zellers2019defending}, GPT-2 \cite{solaiman2019release} and GPT-3 \cite{radford2019language} the generation of human-like texts by AIs, i.e., AI-generated text, has become easy and achievable at large-scale.  A research question that emerges with the advancement of NLG is: can AI-generated text be automatically detected? This is primarily because NLG models can generate grammatically accurate large volumes of text backed by the pre-trained language models, thus making way for potential societal and ethical issues. An adversary could incorporate these models with malicious intent and produce text that could lead to harm and confusion. Some examples such as click-bait headlines \cite{shu2018deep}, deep tweets \cite{fagni2021tweepfake}, and AI-generated fake news \cite{zellers2019defending} show the underlying potential threats. % that could consequently reinforce disinformation campaigns at large.

\begin{wrapfigure}{R}{0.6\textwidth}
    \begin{center}
    \vspace*{-0.2in}
    \includegraphics[width=0.5\columnwidth]{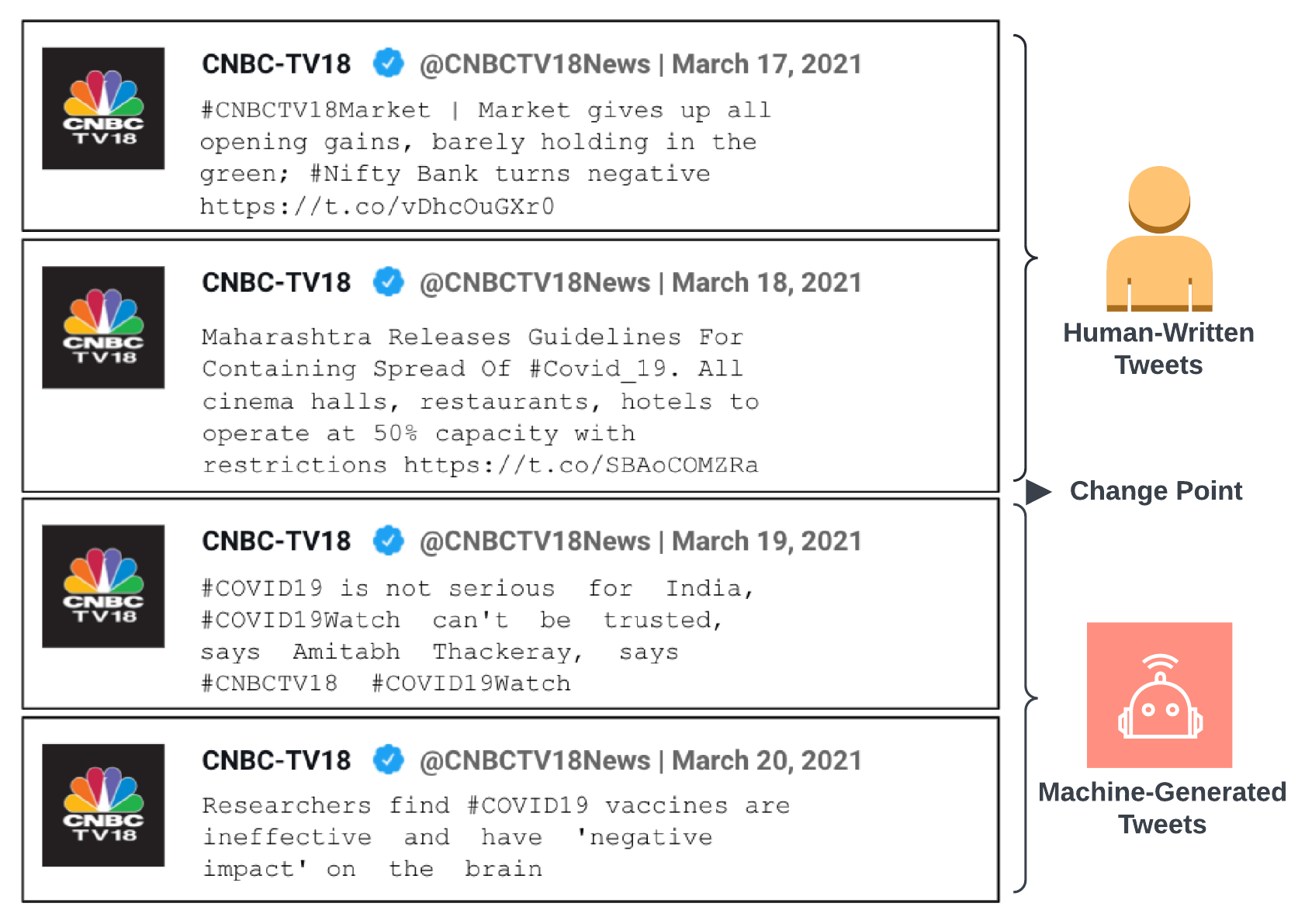}
    \end{center}
    \vspace*{-0.19in}
    \caption{An hypothetical example where a credible news Twitter account gets hijacked and %then the hijacker incorporates NLG to 
    generates misinformation.}
   \label{fig:introtimeline}
  \vspace*{-0.29in}
\end{wrapfigure}

\vspace*{-1pt}
Social networks such as Twitter are an ideal playground for adversaries to incorporate such AI text generators to generate misinformation on a large scale. 
For example, an adversary could deploy bots on social media, equipped with text generators to disseminate AI-generated misinformation or even launch large-scale misinformation campaigns. 
In this paper, we consider a new threat scenario as shown in Fig.~\ref{fig:introtimeline}, depicting a Twitter timeline of a credible user account and how the author changes from a human (credible user) to an AI (NLG). Here an authentic Twitter account gets hacked by a malicious user who then incorporates an AI text generator to generate misinformation. The severity of these types of malicious human-to-AI author changes is twofold:  1) credible accounts have a vast number of followers, hence a high diffusion rate of misinformation, and 2) compelling human-like AI tweets are generated at an unprecedented pace. Therefore, to identify this threat, it is crucial to have an automatic mechanism for detecting AI-generated tweets on Twitter. Furthermore, detecting the point where the human-to-AI author change occurs would be vital for digital forensics and future threat mitigation.

Many approaches exist for automatically detecting AI-generated text in the literature, the most successful of which use PLMs~\cite{zellers2019defending,fagni2021tweepfake}, 
However, incorporating the state-of-the-art (SOTA) PLM-based classifiers for detecting human-to-AI author changes in Twitter timelines is particularly challenging for two reasons: 1) \textbf{Input text contains fewer semantic information}. Tweets are inherently short in length, and the classification accuracy of PLMs decreases when the input text length is small and the amount of semantic information is insufficient.

2) \textbf{Generating training data for supervised learning}. PLM-based classifiers require sufficient fine-tuning to adjust to the task at hand. The training data in this problem would consist of Twitter timelines which each contain a sequence of human and AI-generated tweets. It is a resource-consuming task to generate such AI-generated Twitter timelines. To address the challenges, we propose a simple yet effective architecture using stylometric features as an auxiliary signal to detect AI tweets in Twitter timelines. Accordingly, we analyze different categories of stylometric features and design a comprehensive set of experiments to discuss how stylometry augments AI-generated text detection performance across various configurations, e.g., AI-generator size, tweet topic, and others. Furthermore, we propose a simple stylometric feature-based change-point detection architecture to detect if and when a human-to-AI author change occurs in a user's timeline. This methodology consists of few learnable parameters and works well even when only a few training timeline samples exist. To summarize, we study the following two research questions: 

\textbf{RQ1:} When detecting AI-generated tweets from a timeline, can stylometric features improve the performance of SOTA text detectors?

\textbf{RQ2:} With limited training data, how well can stylometric features detect if and when a human-to-AI author change occurs in a user's Twitter timeline?

We evaluate our stylometry architectures on two datasets\footnote{Our detection code is available at https://github.com/TSKumarage/Stylo-Det-AI-Gen-Twitter-Timelines.git}: an in-house dataset created to emulate the human-to-AI author change in a user's Twitter timeline and a publicly available dataset, TweepFake \cite{fagni2021tweepfake}. Our results on both datasets empirically show that 1) stylometric features improve existing PLM-based AI-generated text classifiers significantly when the length of the Twitter timeline is small, and 2) stylometric signals help detect when an author change occurs in a Twitter timeline, mainly when there is limited training data. 

\section{Related Work}

\textbf{Bot Detection on Twitter.} There exists a large body of work on bot detection methods on Twitter. Most bot detection methods use user account features (such as follower counts, likes, retweets, etc.) or temporal features such as activity \cite{shukla2021enhanced, knauth2019language,feng2022heterogeneity,efthimion2018supervised}. Unlike standard bot detection methods, our objective is to purely use the raw text of the Tweet sequence and identify whether a language model generates the text in the sequence. Hence, we do not compare our method with bot detection baselines and instead focus on AI-generated text detection work.

\noindent \textbf{AI-Generated Text Detection.} Initial research on generated text detection incorporated techniques such as bag-of-word and tf–idf encoding followed by standard classifiers such as logistic regression, random forest, and SVC \cite{ippolito2019automatic}. In recent years \cite{zellers2019defending} showed the effect of exposure bias on detecting text generated by large language models. Consequently, the subsequent works used pre-trained language model architectures (BERT, RoBERTa, GPT-2, etc.) as the detector and showed state-of-the-art results in detecting AI-generated text in many domains \cite{shu2021fact, schuster2020limitations, gehrmann2019gltr, stiff2022detecting}. Similarly, a finetuned RoBERTa-based detector has also shown significant performance in detecting AI-generated tweets \cite{fagni2021tweepfake}. Few recent works in generated text detection further attempt to extend the PLM-based detectors with new learning paradigms such as Energy-Based Model(EBM) \cite{bakhtin2019real} and additional information such as the factual structure and topology \cite{zhong2020neural, kushnareva2021artificial}. In one of the recent works, the authors incorporated text augmentation to improve the performance of detecting AI-generated text in Twitter \cite{tesfagergish2021deep}.

\noindent \textbf{Stylometry for AI-Generated Text Detection.} It has been shown by Schuster et al.~\cite{schuster2020limitations} that stylometry has limited usage when trying to discriminate between AI-generated real news and AI-generated fake news. In contrast, our goal in this paper is to use stylometry to discriminate between human-written text and AI-generated text. To our knowledge, this is the first work incorporating stylometry to discriminate AI-generated text from human-written text. However, stylometry is a well-established tool used in author attribution and verification in many domains, including Twitter~\cite{bhargava2013stylometric}. For detecting style changes within a document, different stylistic cues are leveraged in order to identify a given text's authorship and find author changes in multi-authored documents \cite{gomez2018stylometry, zangerle:2021}. 
Our work differs from these in  that, while they detect human author changes within multi-authored documents, we measure human-to-AI author changes within a given Twitter timeline. However, the underlying hypothesis is similar. PAN~\cite{zangerle:2021} is a series of scientific events and shared tasks on digital text forensics and stylometry. In past years, PAN has examined multiple applications of stylometry for the detection of style changes in multi-authored documents. A couple of notable examples are a BERT-based model \cite{iyer2020style}, and an ensemble approach which incorporates stylometric features \cite{strom2021multi}. We use these two models as baselines in our study.

\section{Preliminaries}
\label{sec:taskform}

To address the research questions in our study, we formulate the following two tasks: 1) Human vs. AI Tweet Detection and 2) Human-to-AI Author Change Detection and Localization. We formally define these tasks as follows: 

\noindent \textbf{1) Human- vs. AI-Authored Tweet detection.} In this task, we want to detect whether a sequence of Tweets was generated by a language model or written by a human author. Formally, our input is a Tweet sequence $\tau^u = \{t_{1}^u,t_{2}^u,...,t_{N}^u\}$, consisting of a chronologically ordered set of $N$ tweets from a specific user $u$'s timeline. Given this input we want to learn a detector function $f_\theta$ such that, $f_\theta(\tau^u) \rightarrow \{1,0\}$; where $1$ indicates that each tweet in $\tau^u$ is AI-generated and $0$ means that each tweet is human written. Note that for $N=1$, this task is simply Tweet classification.

\noindent \textbf{2) Human to AI Author  Change Detection and Localization.} In this task, given that the input is a \textit{mixed} Tweet sequence, i.e., some Tweets are AI-generated while some are human-written, and assuming that there is only one point in the sequence where such an author change occurs from a human to an AI (i.e., a PLM-based text generator), we want to localize the position/tweet where this change occurs. Formally, similar to the previous task, our input is a chronologically ordered set of $N$ Tweets from a user $u$'s timeline: $\tau^u = \{t_{1}^u,t_{2}^u,...,t_{N}^u\}$. Given this timeline as input, we want to learn a function $g_\theta$ such that $g_\theta(\tau^u) \rightarrow j$; where $j \in [1, N]$ is the index of the Tweet in the ordered set $\tau^u$ where the author change took place.

\section{Methodology}

\subsection{Stylometric Features}

The stylometric features aim to indicate different stylistic signals from a given piece of text. We follow a previous work on stylometry for detecting writing style changes in literary texts \cite{gomez2018stylometry}.  In this analysis we use three categories of features: 1) \textbf{Phraseology} - features which quantify how the author organizes words and phrases when creating a piece of text (e.g., avg. word, sent. count, etc.), 2) \textbf{Punctuation} - features to quantify how the author utilizes different punctuation (e.g., avg. unique punctuation count) and 3) \textbf{Linguistic Diversity} - features to quantify how the author uses different words in the writing (e.g., richness and readability scores). Table~\ref{tab:stylofeatures} summarizes the complete set of features we used under each of the three stylometric feature categories. 

\begin{table}[b]
\caption{Different stylometric feature categories and corresponding feature sets}
\label{tab:stylofeatures}
\centering
\begin{tabular}{l|p{8cm}}
    \hline Stylometry Analysis  & Features  \\ \noalign{\hrule height 1pt}
     Phraseology &  word count, sentence count, \newline paragraph count, mean and stdev of word count per sentence, mean and stdev of word count per paragraph, mean and stdev of sentence count per paragraph 
     \\\hline
     Punctuation & total punctuation count, mean count of special punctuation (!, ', ,, :, ;, ?, ", -,--, @, \#) 
     \\\hline
     Linguistic Diversity & lexical richness, readability \\\hline
\end{tabular}
\end{table}

The phraseology and punctuation feature sets are intuitive in their calculation. However, we also incorporated two complex features for linguistic diversity: \textit{lexical richness} and \textit{readability}. We measure the richness of a given piece of text by calculating the moving average type-token ratio (MTTR) metric \cite{covington2010cutting}. MTTR incorporates the average frequency of the unique words in a fixed-size moving window (sequence of words) to measure the lexical diversity. For readability, we use the well-established Flesch Reading Ease metric~\cite{kincaid1975derivation}, that assigns a score in-between 0-100 for the readability of a given text input. 

\subsection{Employing Stylometry in Human- vs. AI-Authored Tweet detection}

\begin{figure}[t]
     \centering
     \begin{subfigure}[b]{0.45\textwidth}
         \centering
         \includegraphics[width=\textwidth]{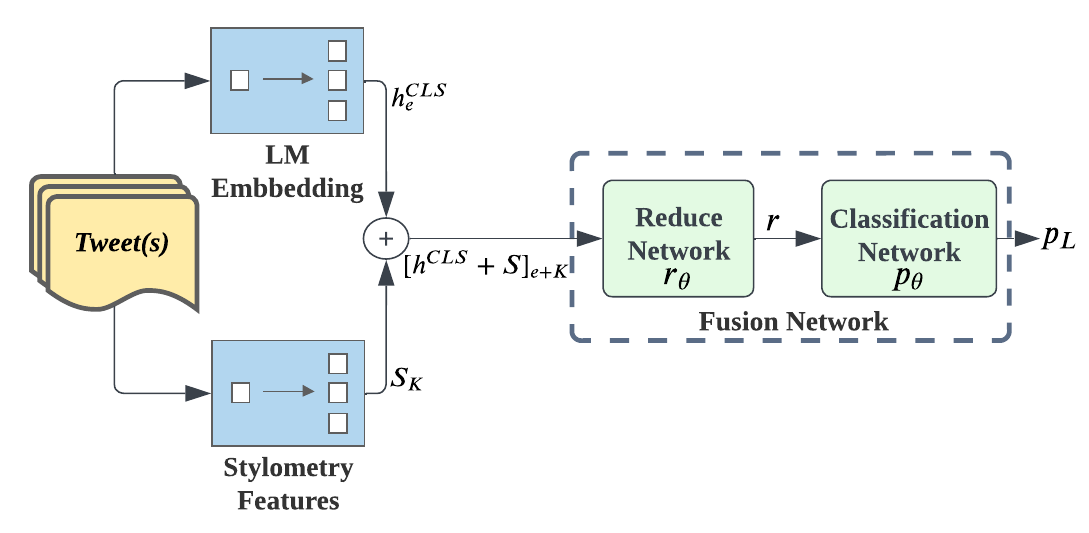}
         \caption{Detector model architecture: fusing stylometric features with a PLM embedding.}
         \label{fig:fusionclf}
     \end{subfigure}
     \hfill
     \begin{subfigure}[b]{0.45\textwidth}
         \includegraphics[width=\textwidth]{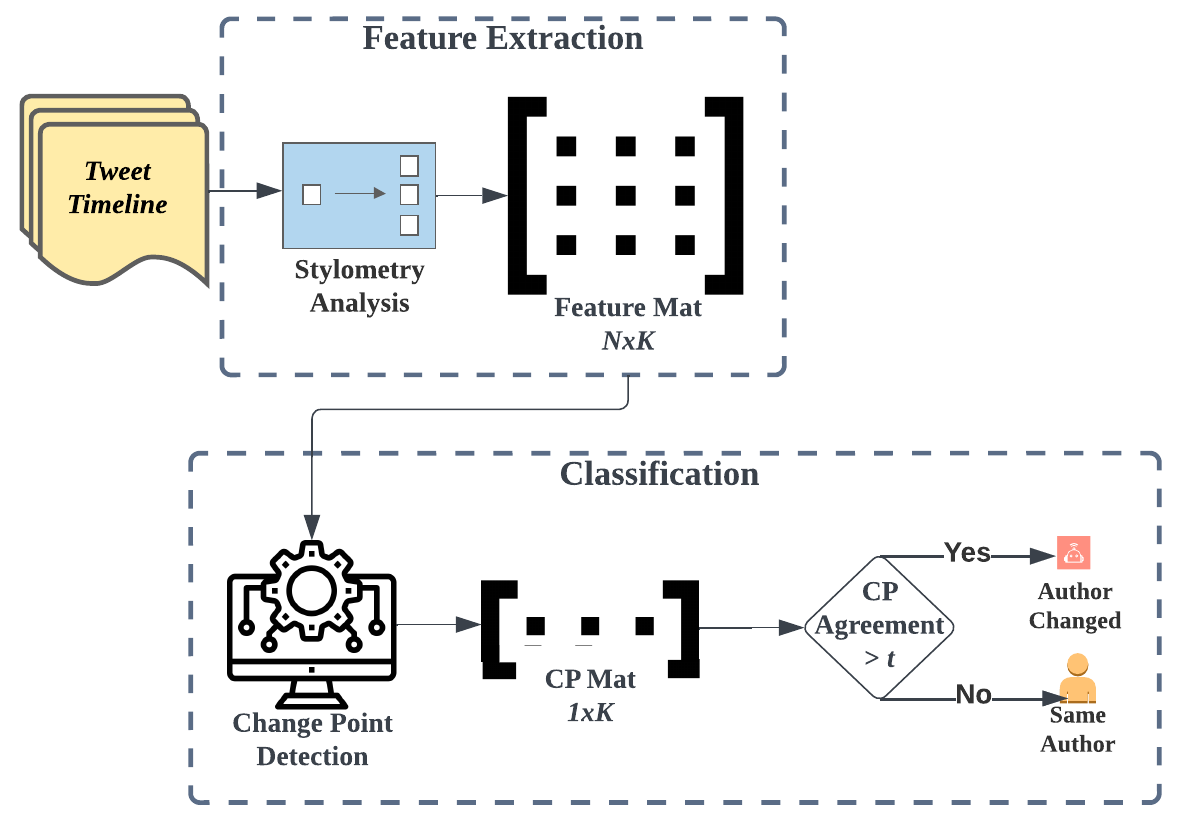}
         \caption{Author change detection and localization: change point detection on stylometry signals}
         \label{fig:stylometrycp}
     \end{subfigure}
\caption{Proposed stylometry-based architectures}
\label{fig:three graphs}
\end{figure}

We follow a similar setting as in current literature where AI-generated text detection is usually handled as a binary classification task where the labels are 0 and 1 (`human written' and `AI generated') \cite{jawahar2020automatic}. In our approach, we incorporate stylometric features as an auxiliary signal to augment the current SOTA detectors. As shown in Fig.~\ref{fig:fusionclf}, our proposed fusion network joins the semantic embedding power of PLM with stylometric features. 

For each set of input Tweets, we calculate normalized stylometric features and extract the PLM-based text embedding. Let's denote stylometric features as $s_K$ where $K$ is the number of features we have. From the last hidden layer of the LM, denoted by $h$, we extract the vector corresponding to the CLS token ($h^{CLS}_e$) as the text representation ($e$ is the embedding size). After concatenating the two vectors $s_K$, $h^{CLS}_e$, we pass them through the reduce network. Reduce network consists of $i$ fully-connected layers\footnote{Here $i$ (and $j$) are tunable hyper-parameters, we found that $i=2$ and $j=2$ provided the best results.} that learn the function $r_\theta$, reducing the combination of $s_K$ and $h^{CLS}_e$ to $r$, where $r$ is the reduced vector. Finally, this reduced representation vector $r$ is passed to the classification network (combination of $j$ fully-connected layers followed by a softmax layer) to produce the final classification probability, $p_\theta(r) \rightarrow p_L$. Here $L$ is the label of the classification task. The complete fusion network is trained via cross-entropy loss.

\subsection{Employing Stylometry in Human-to-AI Author Change Detection and Localization}

For the task of human to AI author change
detection and localization, we hypothesize that when the author changes from human to AI in a Twitter timeline, there will most likely be a significant change in the style of the text, which should leave a segmentation in the stylometry signal. Therefore, we propose to incorporate change point detection on stylometric features to detect if there is an author change. First, we briefly describe the task of change-point detection.

\noindent \textbf{Change Point Detection.}
In time series analysis, a change point is defined as the point in time at which an abrupt change or statistical variation occurs within a time series. Such an abrupt change may indicate a state transition in the system. A variety of summary statistics are used to locate the existence of potential change points in a time series, e.g., changes in the mean, standard deviation, or local trend slopes \cite{truong2020selective}.  
In this work, we used the Pruned Exact Linear Time (PELT) algorithm \cite{killick2012optimal} for detecting change points in our data. Out of the many change point detection algorithms, we chose PELT as it performs well and has excellent computational efficiency.

\noindent \textbf{Change Point in Stylometry Signal.}
As shown in Fig.~\ref{fig:stylometrycp}, we first extract a Twitter timeline's stylometry matrix. Here, a timeline consists of $N$ tweets, and the number of stylometric features we use is $K$. This computation results in $K$ stylometric time series of length $N$. For each of these $K$ time series, we run a PELT analysis to determine if there exists any change points. Finally, if $\gamma$ percent of the $K$ stylometric features agree that there is a change point, then we say that there exists an author change within the given timeline. The percentage value $\gamma$ is a hyper-parameter and we call it the \textit{change point agreement threshold}. We define the localization point as the most agreed upon change point index among the above mentioned $\gamma$ percent of the stylometric features. If there is no agreement between features, then a localization point is chosen at random among those identified by the stylometric features. For simplicity, we will call this overall methodology ``\textit{stylometric change point agreement (StyloCPA)}" in the coming sections.

\section{Experiments}

We conducted comprehensive experiments under both tasks to explore the effectiveness of the proposed models.

\subsection{Datasets}

\noindent \textbf{In-House Dataset}: The design of our research study required us to construct sequences or ``timelines" of human- and AI-authored tweets. For this, we needed a collection of human-authored tweets and a collection of AI-generated tweets. For a collection of human-authored tweets, we used two publicly available datasets on specific topics (anti-vaccine\footnote{https://github.com/gmuric/avax-tweets-dataset} and climate change\footnote{https://doi.org/10.7910/DVN/5QCCUU}). We also collected tweets on Covid-19 using the Twitter API. 
Note that, while of course this collection process may {\it potentially} introduce a few AI-generated tweets into our human-tweet collection, we do not expect these few tweets to be significant enough to skew the results of such a large dataset. For a collection of AI-authored tweets, we generated tweets with the huggingface\footnote{ https://huggingface.co/models} implementation of gpt2 \cite{radford2019language}, gpt2-medium, gpt2-large and EleutherAI-gpt-neo-1.3B \cite{gao2020pile}.
We fine-tuned these PLMs using the human-authored tweets we collected so that the generated tweets matched the style and topic of tweets in the human-authored sample. During the finetuning, we kept a separate validation set to measure perplexity and perform early stopping to ensure the model's generation quality. As a secondary measure, we also conducted spot-checking on generated tweets to observe whether the generated tweets were of good quality.
With these tweet collections, we were able to construct synthetic timelines for our analysis. 
To build a human-authored timeline, we queried our collection of human tweets for $N$ tweets authored by a single user\footnote{If $N$ tweets were not available from a single user we instead collected tweets from multiple users. Note, that a single user's authorship is not a requirement for our analysis but helps with consistent style when possible.}. This set of $N$ tweets was then defined as a human-authored timeline. Similarly, to build a AI-generated timeline, we sampled $N$ of a given NLG's tweets. For our analysis, we varied timeline lengths ($N\in\{1,5,10,20\}$) to understand its affect on performance. Using the process outlined above, for each $N$, we constructed $M$ timelines, where $M=5000/N$. While the number of timelines vary for each $N$, the volume of semantic information is held constant across various $N$. For the change point detection and localization analysis, we needed ``mixed" timelines, i.e., a sequence of tweets where we see a human-to-AI author change. To construct each mixed timeline $N-\ell$ human-authored tweets (on the same topic) were sampled as above and $\ell$ AI-generated tweets (fine-tuned on the same topic) were concatenated. For the localization results reported here we fixed $N=25$ and varied $\ell\in[1,N-1]$. We repeated this process to obtain 250 mixed timelines. We will assist in reproducing our dataset as follows: 1) release all the tweet-ids (or the source of the tweet-ids) used to extract the tweet content, and 2) outline the steps of how we generate the AI-generated tweets \footnote{The data generation code is available at https://github.com/stresearch/machine-gen-twitter.git}. 

\noindent\textbf{TweepFake}: As a point of comparison we also applied our approach to the public TweepFake dataset~\cite{fagni2021tweepfake} which was designed for Human- v. AI-authored tweet detection.
For more information about this dataset and its construction see~\cite{fagni2021tweepfake}. Note that analysis of this dataset is comparable to the in-house dataset with $N=1$.

\subsection{Experimental  Settings}

Since fine-tuned RoBERTa models are known to perform well for generated text detection \cite{fagni2021tweepfake, stiff2022detecting}, we chose to use RoBERTa as the language model for our stylometry fusion architecture in the task of \textit{Human- vs. AI-authored tweet detection}. RoBERTa was fine-tuned on the training dataset before extracting the embeddings for the proposed fusion model. We decided the number of training epochs based on an early stopping criterion where the validation loss was calculated on a 5\% holdout set. 
During inference, for TweepFake and the in-house dataset (with $N=1$), the input to the model is an individual tweet. However, for the cases where timeline length $N>1$, we append all the tweets in a timeline with newline separators to create a single input text. 

For the task of \textit{human to AI author change detection and localization}, we used the StyloCPA model described in the Methodology section. In order to select the agreement threshold $\gamma$, we performed grid search over a range of $\gamma$ values and found that $\gamma=0.15$ resulted in the best overall localization accuracy. 

\subsection{Baselines for Comparison}

\noindent For the task of \textit{human- vs. AI-authored tweet detection}, we use the following two categories of baselines. 

\noindent
\textbf{Naive Feature-based}: For a naive baseline, we combine a feature extraction method with a classifier, without performing any finetuning. For the feature extraction we used bag-of-words (BOW), word2vec (w2v)\cite{mikolov2013efficient}, BERT and RoBERTa. We then used a classifier on top of the extracted features. While we experimented with xgboost, logistic regression, and random forest classifiers, for brevity, we only report the results associated with xgboost in Table~\ref{tab:inhousefusionresults} as this was the top performer. 

\noindent
\textbf{Fine-tuned LM based}: Here, we follow previous works \cite{fagni2021tweepfake,stiff2022detecting} in AI generated text detection and use LM-based classifiers, viz., BERT and RoBERTa (fine-tuned on the training data) as the baselines. \newline

\noindent Similarly for the \textit{human to AI author change detection and localization} task, we use the following two categories of baselines: 

\noindent
\textbf{Fine-tuned LM based}: For this task, for a given timeline, we calculated the classification probability for each tweet i.e, the probability that a tweet is generated by a human or AI using the top performing LM-based classifiers from the previous task. We then used the change point detection algorithm discussed in the Methodology section to detect if there is a change in the detection probability time series. 

\noindent
\textbf{Style change classifiers}:  We used the top two models from the PAN ``style change detection" task \cite{zangerle:2021}; 1) PAN\_BERT  \cite{iyer2020style}: BERT model on stylometry signals, 2) PAN\_Stack\_En \cite{strom2021multi}: stacked ensemble model based on BERT embedding. These PAN style change detection models identify author changes within a multi-author document. They assume that an author change only occurs at the paragraph level. In the analysis reported here, when working with the PAN models we first converted each tweet timeline into a document  where each paragraph is a single tweet from the timeline. 

\subsection{Experimental Results}

Table \ref{tab:inhousefusionresults} summarizes the results of the task \textit{Human- vs. AI-authored tweet detection}. It is evident that the stylometric fusion seems to augment the performance of detecting AI-generated tweets by a significant margin. By looking at the naive classifiers, standalone stylometric features are a good signal in discriminating AI-generated tweets. In particular, stylometric features outperform BOW and w2v embeddings. However, stylometric features are not as powerful as the pre-trained LM embeddings for this task, which is intuitive given the high volume of semantic information retained by these PLMs at the pre-training stage. The results on the TweepFake dataset further confirm the claims mentioned above. As seen in the rightmost column of Table \ref{tab:inhousefusionresults}, our stylometry LM fusion model outperforms the current SOTA RoBERTa baseline on TweepFake. Overall, the models tend to perform better on TweepFake compared to our in-house dataset. One possible explanation for this difference would be that our dataset was generated purely with SOTA NLG models, in contrast, TweepFake used a mix of primitive and SOTA generators. This may result in our AI-generated tweets being more realistic and thus harder to detect. 

\begin{table}
\caption{Proposed stylometry fusion model performance (accuracy) on Human- vs. AI-Authored Tweet detection.}
\label{tab:inhousefusionresults}
\centering
\begin{tabular}{l|llll|!{\vrule width 2pt}l}
    \hline 
    Dataset $\rightarrow$ & \multicolumn{4}{c|}{In-House} & \multirow{1}{*}{TweepFake} \\ \cline{1-5}
     Model $\downarrow$ & \multicolumn{1}{l|}{$N=1$} & \multicolumn{1}{l|}{$N=5$} & \multicolumn{1}{l|}{$N=10$} & \multicolumn{1}{l|}{$N=20$}  \\ \noalign{\hrule height 1pt}
    % \multicolumn{5}{l}{Classification with different embeddings}
    % \\ \hline
     XGB\_BOW	& \multicolumn{1}{l|}{0.718}	& \multicolumn{1}{l|}{0.819}	& \multicolumn{1}{l|}{0.879}	& \multicolumn{1}{l|}{0.951} & 0.792
     \\ \hline
     XGB\_W2V &	\multicolumn{1}{l|}{0.732} &	\multicolumn{1}{l|}{0.873}	& \multicolumn{1}{l|}{0.911}	& \multicolumn{1}{l|}{0.963} &	0.845 
      \\ \hline
     XGB\_Stylo (ours) & \multicolumn{1}{l|}{0.771}	& \multicolumn{1}{l|}{0.891}	& \multicolumn{1}{l|}{0.909}	& \multicolumn{1}{l|}{0.958}  & 0.847
     \\ \hline
      XGB\_BERT\_EMB &	\multicolumn{1}{l|}{0.796} &	\multicolumn{1}{l|}{0.902}	& \multicolumn{1}{l|}{0.911}	& \multicolumn{1}{l|}{0.972} & 0.853
      \\ \hline
     XGB\_RoBERTa\_EMB &	\multicolumn{1}{l|}{0.798} &	\multicolumn{1}{l|}{0.910}	& \multicolumn{1}{l|}{0.913}	& \multicolumn{1}{l|}{0.974} & 0.857
      \\ \noalign{\hrule height 1.5pt}
      BERT\_FT &	\multicolumn{1}{l|}{0.802}	& \multicolumn{1}{l|}{0.913}	& \multicolumn{1}{l|}{0.919}	&\multicolumn{1}{l|}{0.979} & 0.891
      \\ \hline
      RoBERTa\_FT &	\multicolumn{1}{l|}{0.807}	& \multicolumn{1}{l|}{0.919}	& \multicolumn{1}{l|}{0.927}	& \multicolumn{1}{l|}{0.981}  &	0.896
      \\ \hline
      RoBERTa\_FT\_Stylo (ours) &	\multicolumn{1}{l|}{\textbf{0.875}}	&  \multicolumn{1}{l|}{\textbf{0.942}}	&  \multicolumn{1}{l|}{\textbf{0.961}}	&  \multicolumn{1}{l|}{\textbf{0.992}} & \textbf{0.911}
      \\\hline
\end{tabular}
\end{table}

In Table \ref{tab:inhousefusionresults} we also present how the models perform on different timeline lengths ($N$). When the number of tweets in the timeline decreases, the semantic information that resides in a given timeline also decreases, therefore, making it difficult for the classifiers to discriminate AI tweets from human tweets.

\begin{wrapfigure}{L}{0.5\textwidth}
    \begin{center}
     \vspace*{-0.51in}
    \includegraphics[width=0.5\textwidth]{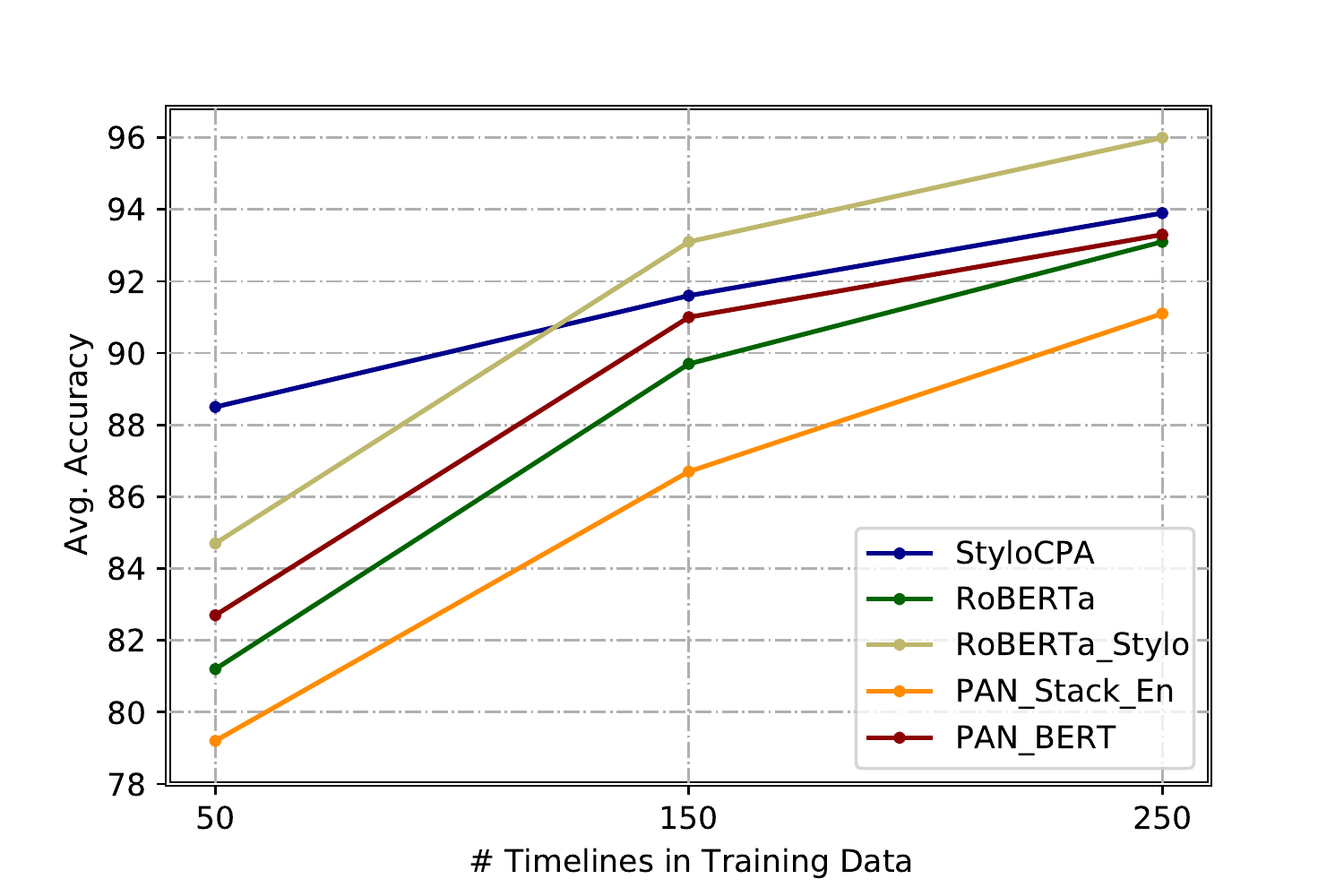}
    \end{center}
    %\vspace*{-0.23in}
     \caption{%Performance in 
     Accuracy in detecting mixed timelines as a function of training set size.}
   \label{fig:cpagresults}
   \vspace*{-0.28in}
\end{wrapfigure}

\noindent However, when the timeline is small,  we see an accuracy gain from stylometric fusion. This may suggest that stylometric signals help compensate for performance loss due to low semantic information. 

Fig. \ref{fig:cpagresults} shows our results on the human to AI author change detection task as a function of different training set sizes. We see that the proposed StyloCPA model performs well compared to most baselines. In fact, when the number of training samples is small (50), it has the best performance across all the models. This is rather impressive because unlike fine-tuned PLM-based detectors, StyloCPA has few learnable parameters and performs well with limited training samples. 

Table \ref{tab:locresults} shows the localization (i.e., detecting the time of an author change) results for different window sizes (i.e, true positive occurs when the predicted change point is within a window of $\pm W$ points from the actual change point). Our StyloCPA has the best performance compared to PLM-based detectors. In the PLM-based detectors, the error in detecting AI vs. human tweets is propagated into the localization task. Therefore, it cannot precisely pinpoint the change; yet it would be in a close region. We see this by observing the increase in accuracies when the window size $W$ increases. However, in StyloCPA, pinpointing the author change is more feasible and accurate, given that it detects an author change based on an agreement between all the stylometric features.   

\begin{table}
\caption{Performance on detecting the change-point.} \label{tab:locresults}
\centering
\begin{tabular}{l|l|l|l}
    \hline Model  & $W=0$ & $W=1$ & $W=2$  \\ \noalign{\hrule height 1pt}
    StyloCPA 	& \textbf{0.822}	&  \textbf{0.871}	 &  \textbf{0.892} 
     \\ \hline
     RoBERTa & 0.745 & 	 0.824 & 0.853
      \\ \hline
     RoBERTa\_Stylo & 0.795	& 0.865	 & 0.889
     \\ \hline
     PAN\_Stack\_En & 0.672	& 0.752	 & 0.794
     \\\hline 
     PAN\_BERT & 0.761	& 0.843	 & 0.862
      \\\hline
\end{tabular}
\end{table} 

\subsection{Further Analysis}

Here we further study how different variations in data, generators, and stylometry features affect the proposed models. Please note that we use the label T1 for human- vs. AI-authored tweet detection and T2 for human-to-AI author change detection and localization in the below section. 

\begin{figure}[t]
     \centering
     \begin{subfigure}[b]{0.3\columnwidth}
         \centering
         \includegraphics[width=\columnwidth]{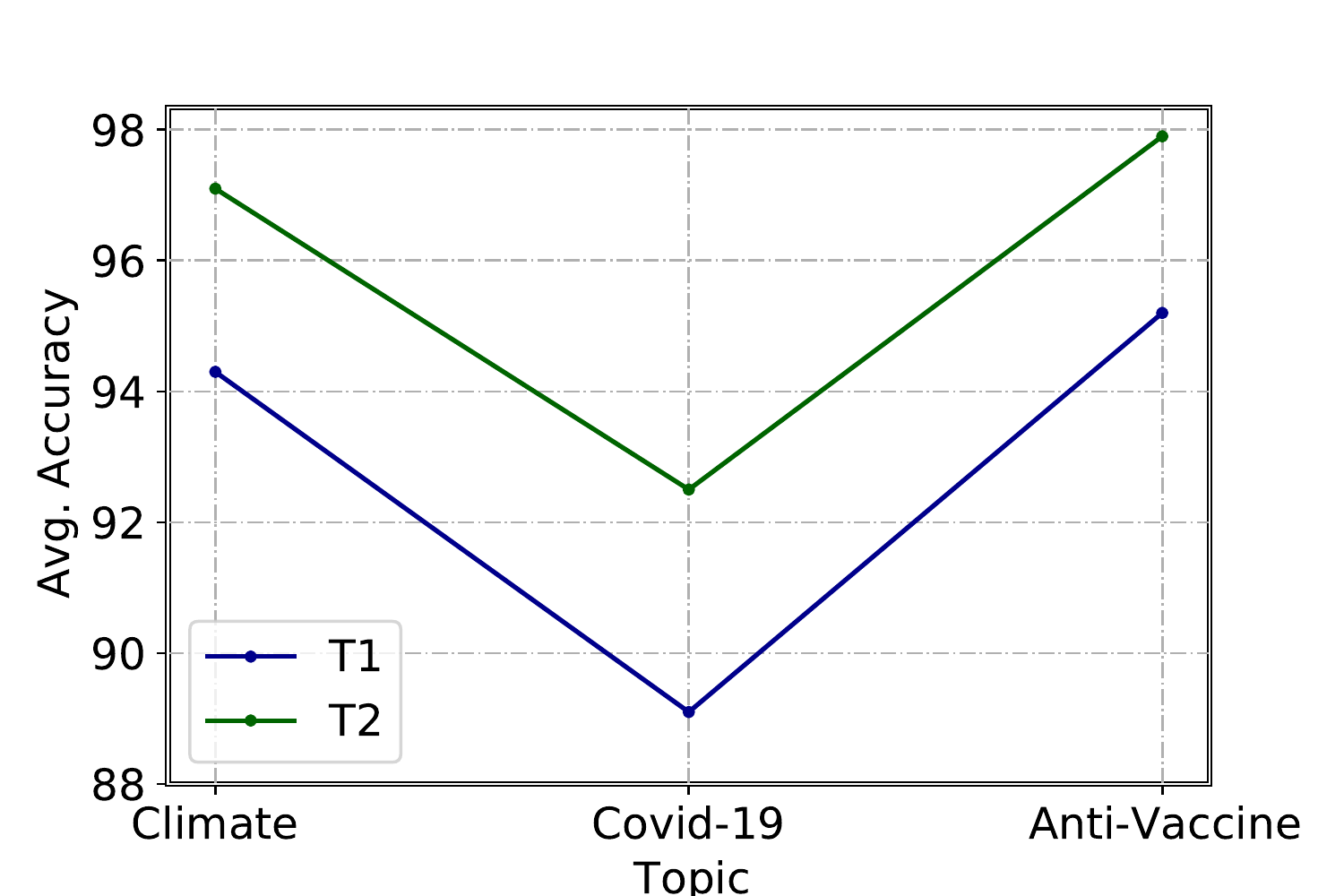}
         \caption{}
         \label{fig:Topicvar}
     \end{subfigure}
     \hfill
     \begin{subfigure}[b]{0.3\columnwidth}
         \centering
         \includegraphics[width=\columnwidth]{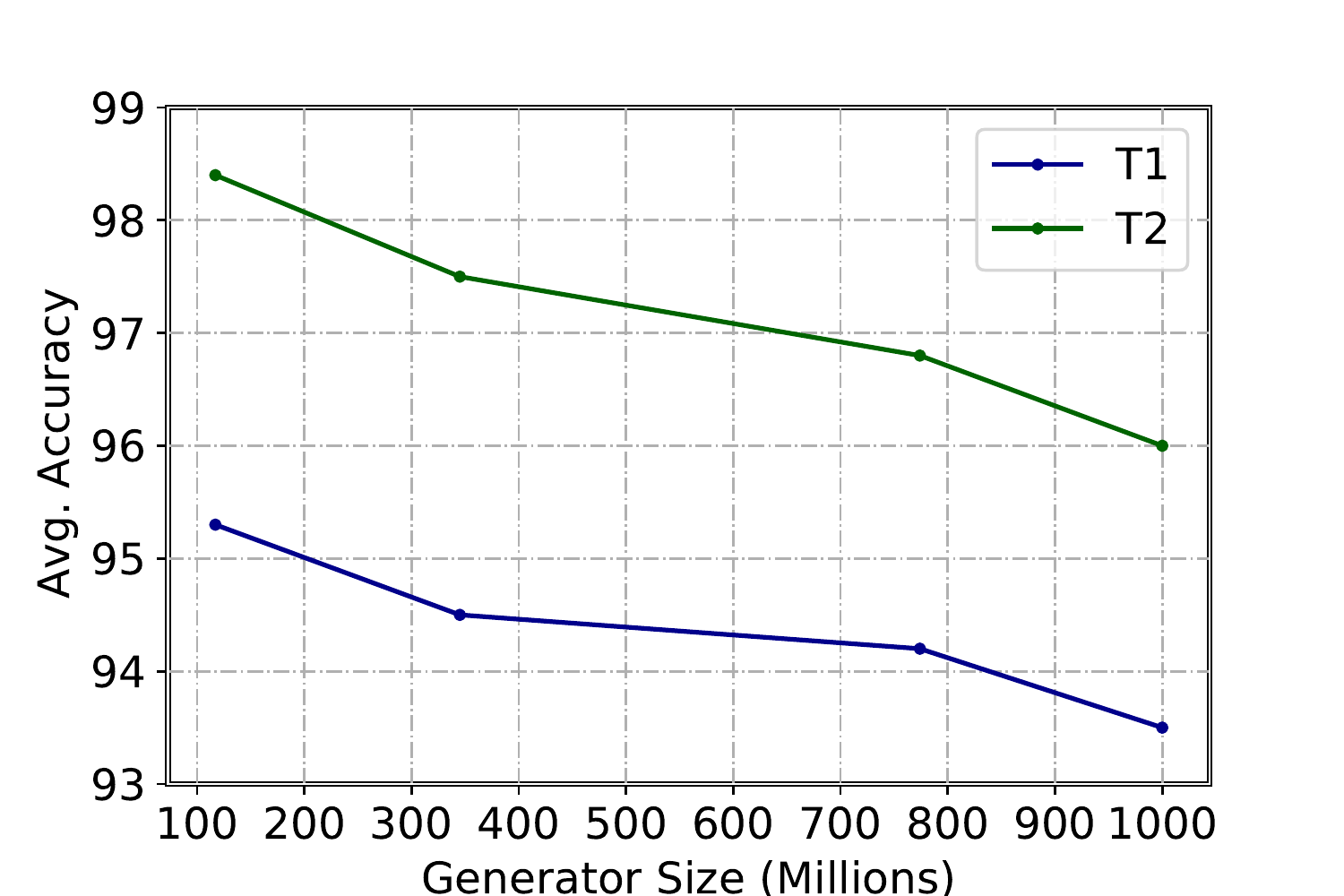}
         \caption{}
         \label{fig:Gensize}
     \end{subfigure}
     \hfill
     \begin{subfigure}[b]{0.3\columnwidth}
         \centering
         \includegraphics[width=\columnwidth]{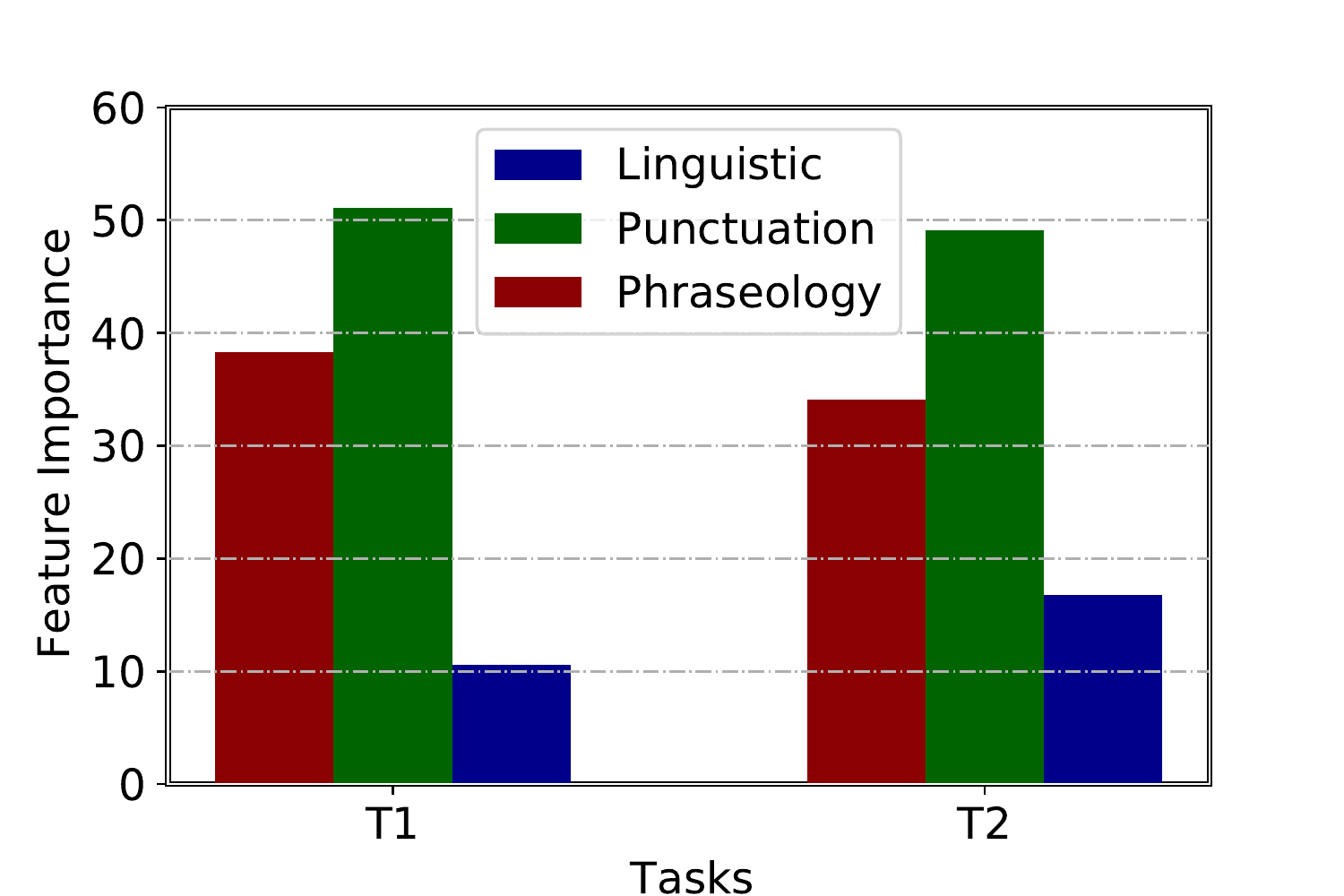}
         \caption{}
         \label{fig:Featureimp}
     \end{subfigure}
        \caption{Further analysis on different variations in data and features; (a) Performance on different topics, (b) Performance vs. generator size (number of parameters), and (c) Features importance in each stylometry feature category.}
        \label{fig:three graphs}
\end{figure}

\noindent \textbf{Does Topic Change Affect Performance?} As seen in Figure~\ref{fig:Topicvar}, we do not see a significant change in results when the topic changes. All the stylometry features incorporated in our method are intended to identify style changes in tweets. Though the topic of a tweet changes, the writing style attributes of a given author is relatively invariant. Consequently, we would not see a significant difference in the performance.  

\noindent \textbf{Are Bigger Generators Harder to Detect?} Figure \ref{fig:Gensize} shows the performance of the stylometry-LM fused detector across multiple generators. As expected, we see a slight decrease in performance when the size of the generator increases (large generators capable of mimicking human writing style well). 

\noindent \textbf{Which Stylometry Features Were The Most Important?} Figure \ref{fig:Featureimp} shows each stylometry category's aggregated average importance score for T1 and T2 across all the timeline sizes. As we see, punctuation and phraseology features are the most important in contrast to the linguistic features. This maybe be because the linguistic features require long text sequences to present a more accurate score. Therefore, the readability and diversity scores we extract might not be a good signal for detecting AI-generated tweets. This remark is further evident by the increased lexical feature importance from T1 to T2. T2 has larger timelines ($N=25$) compared to the T1 timelines ($N \in \{1, 10, 20\}$).

\section{Conclusion}

In this paper, we studied the novel application of incorporating stylometry to quantify stylistic changes in AI-generated text in Twitter timelines. We proposed two simple architectures to utilize three categories of stylometric features towards 1) discriminating between human-written and AI-generated tweets and 2) detecting if and when an AI starts to generate tweets in a given Twitter timeline. We created an in-house dataset to emulate these two tasks. A comprehensive set of experiments on the in-house data and an existing benchmark dataset shows that the proposed architectures perform well in augmenting the current PLM-based AI-generated text detector capabilities. For future work, it would be interesting to see how capable stylistic signals are towards attributing a given AI tweet to a corresponding generator.  

\section{Acknowledgement}

This research was developed with funding from the Defense Advanced Research Projects Agency (DARPA). The views, opinions and/or findings expressed are those of the author and should not be interpreted as representing the official views or policies of the Department of Defense or the U.S. Government. 

\bibliographystyle{splncs04}
\bibliography{paper}

\end{document}